\documentclass[nohyperref]{article}

\usepackage{microtype}
\usepackage{graphicx}
\usepackage{caption}
\usepackage{booktabs} 
\usepackage{xcolor}
\usepackage{csquotes}
\usepackage{amsmath}
\usepackage{amssymb}
\usepackage{mathtools}
\usepackage{amsthm}
\usepackage{multirow}
\usepackage{placeins}
\usepackage{tablefootnote}
\usepackage{hyperref}
\usepackage{url}

\renewcommand\vec[1]{\mathbf{#1}}

\newcommand\vecq[1]{\vec{\widehat{#1}}}

\newcommand\ct[1]{#1}
\renewcommand\vec[1]{\mathbf{#1}}
\newcommand\tlossb[0]{\mathcal{L}}

\newcommand{\rn}[1]{\text{ResNet{#1}}}
\newcommand{\mnv}[1]{\text{MobileNetV{#1}}}
\newcommand{\efnet}[1]{\text{EfficentNet-{#1}}}

\newcommand\eop[1]{\mathop{\mathbb{E}}\left[#1\right]}
\newcommand\eopX[2]{\mathop{\mathbb{E}_{#1}}\left[#2\right]}

\DeclarePairedDelimiter\round{\lfloor}{\rceil}

\DeclarePairedDelimiter\norm{\lVert}{\rVert}

\newcommand{\INT}{\text{int}}
\newcommand\clip[3]{\text{clip}\left(#1,#2,#3\right)}

\newcommand\up[1]{#1_{\uparrow}}
\newcommand\down[1]{#1_{\downarrow}}

\renewcommand{\l}{\left}
\renewcommand{\r}{\right}

\usepackage[accepted]{icml2022}

\usepackage[capitalize,noabbrev]{cleveref}

\theoremstyle{plain}

\theoremstyle{definition}

\theoremstyle{remark}

\usepackage[textsize=tiny]{todonotes}

\icmltitlerunning{Overcoming Oscillations in QAT}

\begin{document}

\twocolumn[
\icmltitle{Overcoming Oscillations in Quantization-Aware Training}



\icmlsetsymbol{equal}{*}

\begin{icmlauthorlist}
\icmlauthor{Markus Nagel}{equal,qualAI}
\icmlauthor{Marios Fournarakis}{equal,qualAI}
\icmlauthor{Yelysei Bondarenko}{qualAI}
\icmlauthor{Tijmen Blankevoort}{qualAI}
\end{icmlauthorlist}

\icmlaffiliation{qualAI}{Qualcomm AI Research, an initiative of Qualcomm Technologies, Inc.}

\icmlcorrespondingauthor{Markus Nagel}{markusn@qti.qualcomm.com}
\icmlcorrespondingauthor{Marios Fournarakis}{mfournar@qti.qualcomm.com}

\icmlkeywords{Machine Learning, ICML}

\vskip 0.3in
]



\printAffiliationsAndNotice{\icmlEqualContribution} 

\begin{abstract}
When training neural networks with simulated quantization, we observe that quantized weights can, rather unexpectedly, oscillate between two grid-points. The importance of this effect and its impact on quantization-aware training (QAT) are not well-understood or investigated in literature. 
In this paper, we delve deeper into the phenomenon of weight oscillations and show that it can lead to a significant accuracy degradation due to wrongly estimated batch-normalization statistics during inference and increased noise during training. These effects are particularly pronounced in low-bit ($\leq$ 4-bits) quantization of efficient networks with depth-wise separable layers, such as MobileNets and EfficientNets. 
In our analysis we investigate several previously proposed QAT algorithms and show that most of these are unable to overcome oscillations. 
Finally, we propose two novel QAT algorithms to overcome oscillations during training: \textit{oscillation dampening} and \textit{iterative weight freezing}. We demonstrate that our algorithms achieve state-of-the-art accuracy for low-bit (3 \& 4 bits) weight and activation quantization of efficient architectures, such as \mnv{2}, \mnv{3}, and \efnet{lite} on ImageNet.
Our source code is available at~\url{https://github.com/qualcomm-ai-research/oscillations-qat}.
\end{abstract}

\vspace{-0.3cm}

\section{Introduction}\label{sec:introduction}
Quantization is one of the most successful methods for optimizing neural networks for efficient inference and on-device execution while maintaining a high accuracy. By compressing the weights and activations from the regular 32-bit floating-point format to more efficient low bit fixed-point representations, such as INT8, we can reduce power consumption and accelerate inference when deploying neural networks on edge devices \citep{horowitz}.


Despite its clear power and latency benefits, quantization comes at the cost of added noise due to the reduced precision. However, researchers in recent years have shown that neural networks are robust to this noise and can be quantized to 8-bits with a minimal drop in accuracy using \textit{post-training quantization} techniques (PTQ) \citep{dfq, bannerposttraining, nahshan2020lapq}. PTQ can be very efficient and, generally, only requires access to a small calibration dataset, but suffers when applied to low-bit quantization ($\leq$ 4-bits) of neural networks. Meanwhile, \textit{quantization-aware training} (QAT) has become the de-facto standard method for achieving low-bit quantization while maintaining near full-precision accuracy \citep{krishnamoorthi,lsq, whitepaper}. By simulating the quantization operation during training or fine-tuning, the network can adapt to the quantization noise and reach better solutions than with PTQ.  
 

In this paper, we focus on the \textit{oscillations} of quantized weights that occur during quantization-aware training. This is a little-known and under-investigated phenomenon in the optimization of quantized neural networks, with significant consequences for the network during and after training. When using the popular \textit{straight-through estimator} (STE) \citep{bengio2013estimating} for QAT, weights seemingly randomly oscillate between adjacent quantization levels leading to detrimental noise during the optimization process. Equipped with this insight, we investigate recent advances in QAT that claim improved performance and assess their effectiveness in addressing this oscillatory behavior.

An adverse symptom of weight oscillations is that they can corrupt the estimated inference statistics of the batch-normalization layer collected during training, leading to poor validation accuracy. We find that this effect is particularly pronounced in low-bit quantization of efficient networks with depth-wise separable layers, such as MobileNets or EfficientNets, but can be addressed effectively by re-estimating the batch-normalization statistics after training. 

While batch-normalization re-estimation overcomes one significant symptom of the oscillations, it does not tackle its root cause. To this end, we propose two novel algorithms that are effective at reducing oscillations: \textit{oscillation dampening} and  \textit{iterative weight freezing}. By addressing oscillations at their source, our methods improve accuracy beyond the level of batch-normalization re-estimation. We show that both methods achieve state-of-the-art results for 4 and 3-bit quantization of efficient networks, such as \mnv{2}, \mnv{3}, and \efnet{lite} on ImageNet.

\section{Oscillations in QAT}\label{sec:motivation}
We first investigate why weights oscillate in quantization-aware training and how this phenomenon affects neural network training in practice.

\subsection{Quantization-aware training}
\label{sec:qat}
One of the most effective ways to quantize a neural network is by training the network with simulated quantization. During the forward pass, floating-point weights and activation are quantized using the quantization function $q(\cdot)$. It takes input vector $\vec{w}$ and returns quantized output $\vecq{w}$ given by:
\begin{equation}
    \label{eq:quantization_formulation}
    \vecq{w} = q(\vec{w}; s, n, p) = \ct{s} \cdot \clip{\round*{\frac{\vec{w}}{\ct{s}}}}{\ct{n}}{\ct{p}}, 
\end{equation}
where $\round*{\cdot}$ is the \textit{round-to-nearest} operator, $\clip{\cdot}{\alpha}{\beta}$ is a clipping function with lower and upper bounds $\alpha$ and $\beta$, respectively, $\ct{s}$ is a scaling factor, and $\ct{n}$ and $\ct{p}$ the lower and upper quantization thresholds.
In this formulation, the quantized weights $\vecq{w}$ are the ones used during inference, whereas the original floating-point weights $\vec{w}$ only serve as proxies for the optimization and are commonly referred to as \textit{latent weights} or \textit{shadow-weights}.

A fundamental challenge in the QAT formulation is that the rounding function in equation \eqref{eq:quantization_formulation} does not have a meaningful gradient, which makes gradient-based training impossible. One of the most popular techniques for alleviating this issue involves using the straight-through estimator (STE) \cite{bengio2013estimating, hinton2012ste} to approximate the true gradient during training. In practice, this means that we approximate the gradient of the rounding operator as 1 within the quantization limits. We can thus define the gradient of the loss $\tlossb$ with respect to $\vec{w}$ as:
\begin{equation}
    \label{eq:dl_dw_ste}
    \frac{\partial\vec{\tlossb}}{\partial \vec{w}} = \frac{\partial\vec{\tlossb}}{\partial \vecq{w}} \cdot \mathbf{1}_{n \leq \vec{w}/\ct{s} \leq p},
\end{equation}
where $\mathbf{1}$ is the indicator function that is $1$ if $\vec{w}$ falls within the quantization grid, and 0 otherwise, such that there is no gradient outside of the representable quantization region. The STE gradient approximation has been widely adopted in recent literature closing the gap between quantized and full-precision accuracy for a wide range of tasks and networks \citep{Gupta2015, dorefa, krishnamoorthi, tqt, lsq, lsq+, whitepaper}.

\begin{figure*}[ht]
\centering
\includegraphics[width=0.98\textwidth]{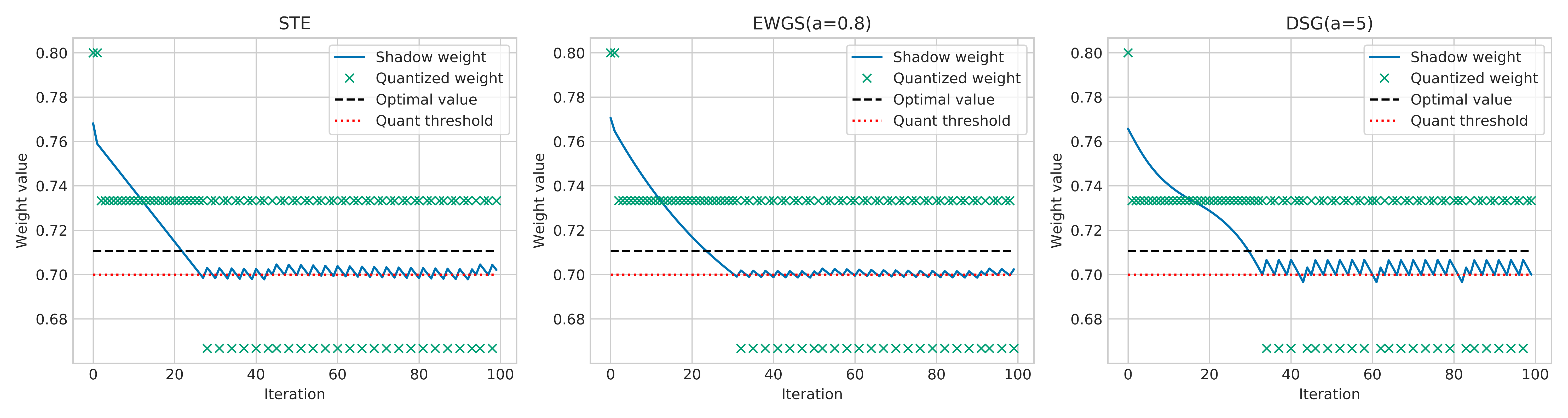} \vspace{-.4cm}
\caption{Oscillation  of a single weight in toy-regression problem for various gradient estimator methods: STE (left), EWGS \citep{EWGS} (middle), and DSQ \citep{DSQ} (right).}
\label{fig:oscillation_single_weight}
\end{figure*}

\subsection{Oscillations}
\label{sec:oscillations}
Despite its wide adaptation and enormous success in QAT, the STE  has a counter-intuitive and very interesting side-effect. The STE causes implicit stochasticity during the optimization process due to the latent weights oscillating around the decision boundary between adjacent quantized states. This phenomenon was recently also observed by \citet{diffq}.

To illustrate this, we propose a simple toy regression example that we will refer to frequently throughout this paper.
We start with an optimal floating-point weight, $w_*$, as target for our 1D toy regression problem and sample a data vector $\vec{x}$ from a distribution $X$ with bounded variance such that, $\eop{\vec{x}^2}=\sigma^2$, where $0<\sigma^2<\infty$. Then we optimize the least-squares problem
\begin{equation}
    \min_{w}\tlossb(w) = \eopX{\vec{x}\sim X}{\frac{1}{2} \left(\vec{x}w_* - \vec{x}q(w)\right)^2},
    \label{eq:toy_regression_optim}
\end{equation}
where $q(\cdot)$ is the quantizer from equation \eqref{eq:quantization_formulation} and $w$ is the latent weight. We optimize the objective using the STE formulation for the gradients. From figure \ref{fig:oscillation_single_weight} (left), we can see that as the latent weight $w$ approaches the optimal value $w_*$, it starts oscillating around the decision threshold between the quantization levels above $\up{w}$ and below $\down{w}$ the optimal value, as opposed to converging to the region closer to the optimal value $q(w_*)$. 

The weight oscillates around the decision threshold due to the fact that the gradient in equation \eqref{eq:dl_dw_ste} is constant above the threshold, pushing the latent weight down towards $\down{w}$, and constant below the threshold pushing the latent weight up towards $\up{w}$. Please refer to appendix \ref{app:toy_example_gradients} for an analytical expression of this gradient. The induced oscillations happen irrespective of the learning rate. In appendix \ref{app:learning_rates_oscillation} we show that decreasing the learning rate reduces the amplitude of the oscillations but do not affect their frequency.  

The frequency of the oscillation is depends on the distance of the optimal value from its closest quantization level,  $d =|w_* - q(w_*) |$. Let's assume that  $q(w_*)=\up{w}$, then the gradient below the threshold is $k=s/d-1$ times larger than the gradient above, where $s$ is scaling factor or quantization step-size and $s>d$. Interpreting the gradient as the velocity of the latent weight, $w$ needs to travel for $k$ iterations before it crosses the threshold and its velocity/gradient is reversed. In appendix \ref{app:relation_distance_frequency} we show experimentally that the oscillation frequency is indeed directly proportional to closeness of $w_*$ to $q(w_*)$. 

It is interesting to note that this behavior bears similarities to stochastic rounding \citep{Gupta2015}, where the closeness of the latent weight to the quantization level is related to the probability of rounding to that level. However, in STE the source of the stochasticity stems from the discrete nature of the gradients and not from sampling. We note that oscillations are not unique to the vanilla STE, but are present in several variations of the STE proposed in literature, a small subset of which we present in figure \ref{fig:oscillation_single_weight}.

 


\begin{figure}[t]
\centering
\vspace{-.2cm}
\includegraphics[width=\linewidth]{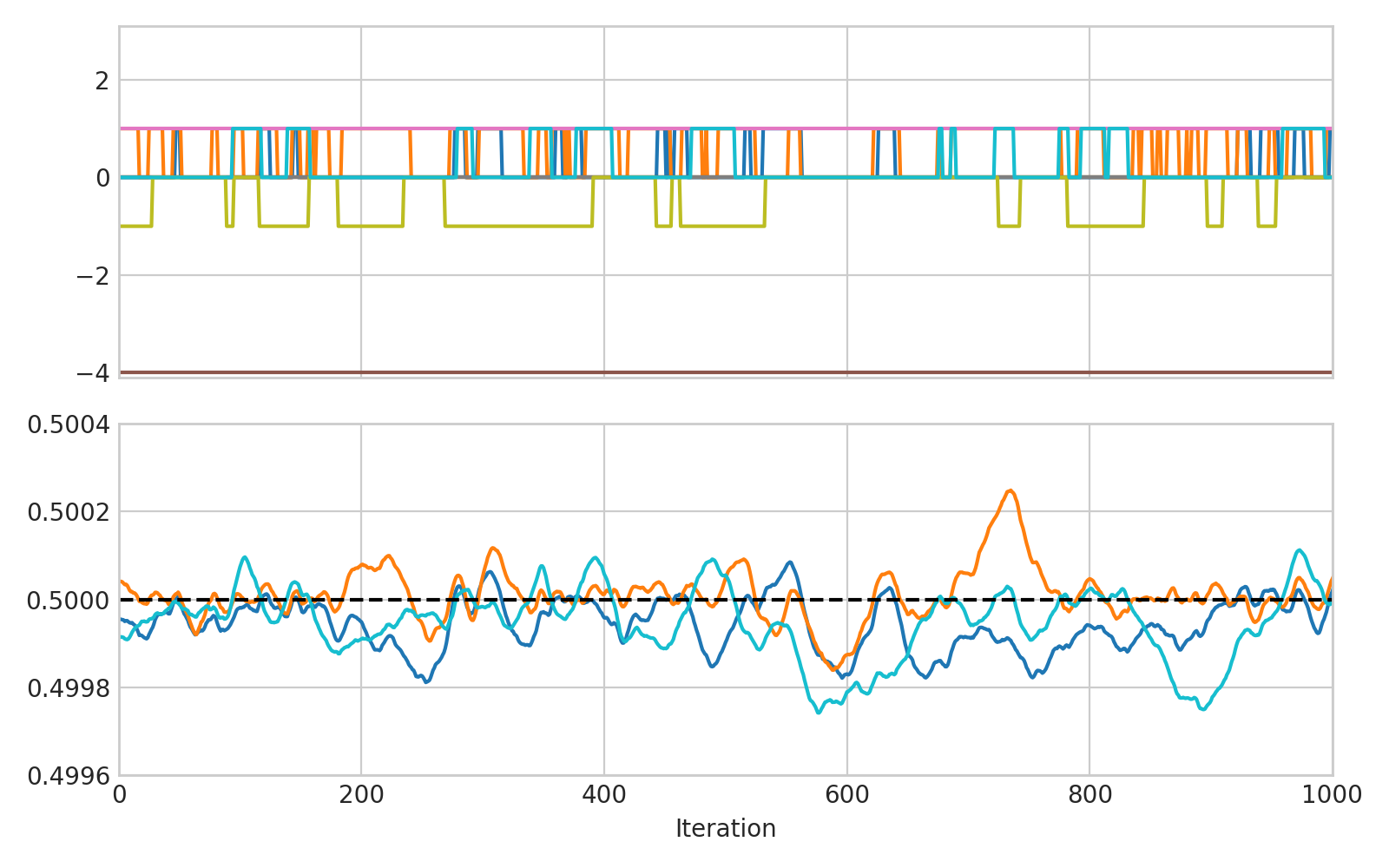} \vspace{-.7cm}
\caption{Progression of weights in the first channel of the first depth-wise separable layer (conv.1.0) during the last 1000 iterations of training \mnv{2} on ImageNet. The top plot shows the integer weights and the bottom plot zooms in on the decision boundary between 0 and 1 for the corresponding latent weights.}
\vspace{-.2cm}
\label{fig:mv2_oscillating_weigths}
\end{figure}

\begin{figure}[t]
\centering
\vspace{-.2cm}
\includegraphics[width=\linewidth]{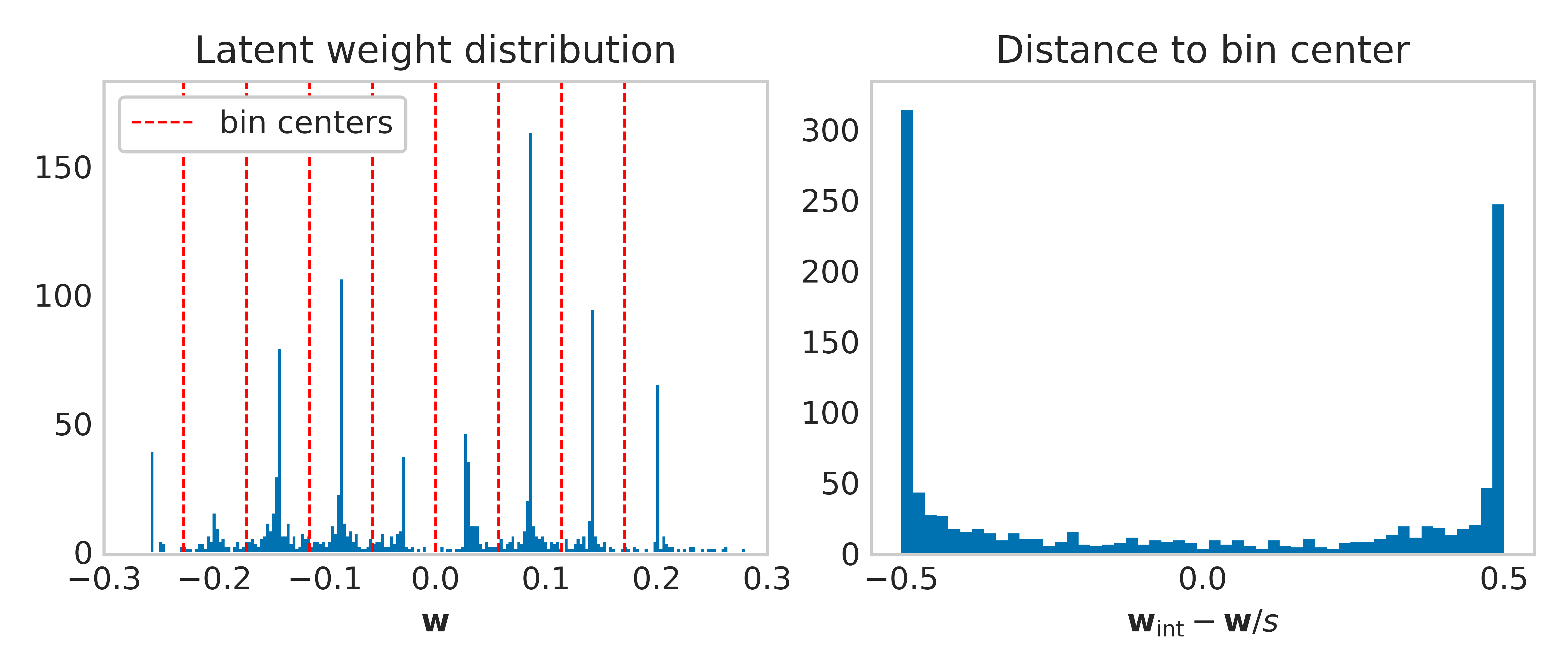} 
\vspace{-.8cm}
\caption{Weight distribution in depth-wise separable layer of third inverted residual block of \mnv{2} (conv.3.1). Histogram of latent weights (left). Histogram of the distance of latent weight from their closest quantization grid-point ($\vec{w}_{\INT} - \vec{w}/\ct{s}$) (right). 
}
\vspace{-.1cm}
\label{fig:mv2_bin_distribution}
\end{figure}

\subsection{Oscillations in practice}
\label{sec:oscillation_in_practice}
These oscillations are not just an unfortunate side-effect of this fictitious toy example. They are indeed present in larger neural networks with significant implications for their optimization. Figure \ref{fig:mv2_oscillating_weigths} shows the progression of 3-bit quantized weights in a depth-wise separable layer of \mnv{2} close to convergence, trained using LSQ \citep{lsq} on ImageNet. We observe that many of the weights appear to randomly oscillate between two adjacent quantization levels. In figure \ref{fig:mv2_bin_distribution}, we also see that after the supposed convergence of the network, a large fraction of the latent weights lie right at the decision boundary between grid points. This further reinforces the observation that a significant proportion of weights oscillate and do not converge. 

We identify two major issues associated with the oscillations in neural network training: Wrong estimation of the batch-normalization \citep{batchnorm} inference statistics, and an adverse effect on the optimization of the network.

\subsubsection{The effect on Batch-Normalization}
\label{sec:effect_on_bn_statistics}

During training  batch-normalization layers track the exponential moving average (EMA) of the mean and variance for each layer's output, so that it can be used during inference as an approximation of the real sample statistics. 
When training full-precision neural networks we can expect that weights will change very slowly close to convergence. As a result, the output statistics of each layer are expected to be quite stable across iterations and therefore the EMA is a good estimate of the statistics.
However, oscillations in QAT can lead to rapid changes of the integer weights (see in figure \ref{fig:mv2_oscillating_weigths}) leading to significant distribution shift between iterations, even close to convergence.
The sudden and large changes in the output distribution induced by oscillations can corrupt the EMA statistics leading to a significant degradation in accuracy. In fact, there are two factors that amplify this effect: the weight bit-width and the number of weights per output channel.

The lower the bit-width $b$, the larger the distance becomes between quantization levels, as it is proportional to $1/2^b$. As the oscillating weights move from one quantization level to the other, they cause a proportionally larger shift in the output distribution. The second important factor is the number of weights per output channel. The smaller the number of weights, the larger the contribution of individual weights to the final accumulation. When the number of accumulations increases, the effects of oscillations average out due to the law of large numbers. 

In table \ref{tbl:bn_stats_diff}, we use the KL-divergence\footnote{We we assume that population $p$ and estimated $q$ outputs are normally distributed, such that $D_\text{KL}(p,q)=\log{\frac{\sigma^2_2}{\sigma^2_1} + \frac{\sigma^2_1+(\mu_1-\mu_2)^2}{2\sigma_2^2} -\frac{1}{2}}$, where $p(x)=\mathcal{N}(\mu_1, \sigma_1)$ and $q(x)=\mathcal{N}(\mu_2, \sigma_2)$.} to quantify the discrepancy between population and estimated statistics. We indeed observe that KL divergence is much larger for depth-wise separable layers than in point-wise convolutions in \mnv{2} and the full convolutions in \rn{18}.
 
\begin{table}[t]
    \centering
    \begin{tabular}{ l | l | r r}
    \toprule
                Network   &   Layer &   $\max{D_{\text{KL}}}$& $
        \eop{D_{\text{KL}}}$ \\
        \midrule

        \rn{18} & layer1.0.conv1 &  0.0059 &	0.0002 \\
        \rn{18} & layer1.0.conv2 &  0.0130 &	0.0014 \\
        \rn{18} & layer3.0.conv1 &  0.0006 & 0.0001 \\
        \midrule
        
        \mnv{2} & conv3.0 (PW)  &  0.7858 &   0.0292 \\
        \mnv{2} & conv3.1 (DW)  &  55.3782 &	1.25464 \\
        \mnv{2} & conv3.2 (PW) &  0.0065 &	0.0012 \\
        \midrule
        \mnv{2} & conv10.0 (PW) &  0.0037	& 0.0004 \\
        \mnv{2} & conv10.1 (DW) &  27.2618 & 0.2900\\
        \mnv{2} & conv10.2 (PW)  &  0.0267 &	0.0034\\
\bottomrule
\end{tabular}
 \caption{Discrepancy between estimated and actual population statistics of batch-normalization following convolutional layers in \rn{18} and \mnv{2} with weights quantized to 3 bits. We report the maximum and mean KL-divergence across the output channels of each layer. DS: depth-wise separable, PW: point-wise.}\vspace{-.1cm}
 \label{tbl:bn_stats_diff}
\end{table}

A cheap and straightforward solution to this problem is to re-estimate the batch-normalization statistics with a small subset of data after training. This method, which we call \textit{batch-normalization (BN) re-estimation}, is sometimes used in stochastic quantization formulations \citep{peters2018probabilistic,louizos2018relaxed}. However, we argue that it is also essential in deterministic QAT formulations due to the oscillating weights. 

In table \ref{tbl:bn_reestimation_diff}, we see that BN re-estimation not only improves the final quantized accuracy for \mnv{2} but also reduces the variance among different seeds. We further observe that the gap in accuracy widens as the bit-width goes down for \mnv{2}, which is not the case for \rn{18}. 





\begin{table}[t]
    \centering
    \begin{tabular}{ l | c | c c }
    \toprule
  Network & Bits &  pre-BN & post-BN \\
        \midrule
\rn{18} & 4 & 70.15\textsuperscript{0.03} & 70.20\textsuperscript{0.02} \\
\rn{18} & 3 & 69.63\textsuperscript{0.01}  & 69.70\textsuperscript{0.05} \\
\midrule
\mnv{2}& 8 & 71.79\textsuperscript{0.07}  & 71.89\textsuperscript{0.05}\\ 
\mnv{2}& 4 & 68.99\textsuperscript{0.44} & 71.01\textsuperscript{0.05}\\ 
\mnv{2}& 3 & 64.97\textsuperscript{1.23} & 69.50\textsuperscript{0.04}\\
\bottomrule
\end{tabular}
 \caption{Validation accuracy (\%) on ImageNet before (pre-BN) and after BN re-estimation (post-BN) for networks trained using low bit weight quantization \citep{lsq} for 20 epochs. We report the average of 3 seeds and the STD in the superscript.}
 \label{tbl:bn_reestimation_diff}
\end{table}



\begin{table}[t]
    \centering
    \begin{tabular}{ l | c  c }
    \toprule
  Method & Train Loss & Val. Acc. (\%) \\
        \midrule
Baseline         & 1.3566  & 69.50 \\
\midrule
SR (mean + std)  & 1.3547\textsuperscript{0.0053}  & 69.58\textsuperscript{0.09} \\
SR (best)        & 1.3391  & 69.85 \\
AdaRound  & 1.3070  & 70.12 \\
\midrule
Freezing & -  & 70.33 \\
\bottomrule
\end{tabular}
 \caption{Ablation study on the effect of oscillations on QAT. Training loss and validation accuracy (\%) for \mnv{2} with 3-bit weight quantization. SR is stochastic rounding of oscillating weights and AdaRound the binary optimization of oscillating weights.}
\vspace{-.1cm}
 \label{tbl:binary_post_opt}
\end{table}
\subsubsection{The effect on training}
Next to harming the BN statistics, oscillations can also have a negative effect on the training process itself. 
To illustrate this we first show that a converged \mnv{2} with 3-bit weights can achieve a lower training loss (and higher validation accuracy) if we randomly sample oscillating weights between their two oscillating states.
To this end, we sample all weights that were oscillating at the end of training with a probability proportional to the time spent at each quantized state, i.e. $p(\up{w}) = \eopX{t}{w^t = \up{w}}$. We calculate the expectation using a exponential moving average over the integer weights as done in line 15 of algorithm \ref{alg:qat_freezing}.

In table \ref{tbl:binary_post_opt}, we present the results of this experiment. 
We observe that the mean training loss over the sampled networks is similar to that of the final converged model.
However, many samples achieve a lower training loss and the best randomly sampled network shows a significantly lower training loss. 
We also perform a binary optimization of the oscillating weights using an adaptation of AdaRound \citep{adaround}. We optimize the rounding of all layers simultaneously on the final task loss, akin to what is done in the literature with simulated annealing to solve binary optimization problems \citep{Kirkpatrick1983SimulatedAnnealing}. We see that this binary optimization significantly improves over the best stochastic sample and the original converged network. This suggests that weight oscillations prevent the network from converging to the best local minimum during training, and can be detrimental to the optimization process. 

Finally, we show that preventing oscillations earlier in training using our oscillation freezing technique (cf. section \ref{sec:freezing_method}) leads to an even higher validation accuracy than the binary optimization of the oscillating weights. This indicates that the oscillations not only prevent QAT from converging to the best local minimum at the end of training, but can also lead the optimizer towards sub-optimal directions earlier in training. 





\section{Related work}\label{sec:backgroundrelated}
\label{sec:related_work}
Earlier work in QAT \citep{guassian_quant, understanding_STE} investigated different variations of the STE and demonstrate experimentally 
that the derivative of the clipped ReLU is a better approximation than the vanilla STE (derivative of identity). In practice, the former is the same as setting the gradient outside the quantization grid to zero, as per equation \eqref{eq:dl_dw_ste}, and is the most widely used formulation in QAT literature \citep{pact2018, lsq, differentiablequantization}.

Despite the success of the STE trick in practice, a lot of prior work has focused on the non-optimality of the STE as a gradient approximator. \citet{diffq} claim that the STE causes instability during training due to weights oscillating leading to \textit{biased} gradients and weights. For this reason, they propose replacing the rounding operation with additive Gaussian noise in the forward pass to simulate quantization noise. Whereas weight oscillations are indeed an important side-effect of the STE, the idea that the STE yields \textit{biased} gradients in a deterministic training setting is opaque, as we discuss in more detail in appendix \ref{app:biased_ste}. \citet{qat_abs_cosine} use a toy example to claim that solutions under STE can lead to a sub-optimal quantized solution but do not recognize that the quantized weight will oscillate around the optimum rather than converge to the wrong value. Based on their observation, they propose an absolute-cosine regularization for the full-precision weights to make them more quantization-friendly but do not compare to other competing QAT methods.

Other work focuses on alternative \textit{gradient estimators} for back-propagation. Despite seeming differences in their motivation and formulations, all these methods in on way or another propose adaptations of STE that boil down into two main categories: \textit{multiplicative} or \textit{additive}. Multiplicative methods apply a scaling to the data-gradient, whereas additive methods add an input-independent term to the gradient through regularization. We provide more details and derivations in appendix \ref{app:toy_example_gradients}. 



\paragraph{Multiplicative} \citet{EWGS} (EWGS) acknowledge that the STE generates \textit{coarse gradients} that treat all latent weights within a quantization bin equally regardless of their distance from its centre and propose scaling the gradient by this distance. However, they do not provide any evidence on why and how these coarse gradient harm optimization.  \citet{PBGS} arrive to a similar formulation but are motivated by robust quantization. They claim that clustering latent weights around bin centres makes them more robust when quantized to different bit widths. In both these methods, the gradient scales linearly with the distance of the latent weight from its quantized value.


\citet{DSQ} (DSQ) try to alleviate the \textit{gradient error} that the STE causes during low-bit quantization but do not provide any evidence about the source or magnitude of this error. They propose using a series of hyperbolic tangent functions to simulate the rounding function in the backward pass. This leads to a non-linear scaling of the STE gradient, with the gradient increasing with the distance from the bin centre. \citet{quantization_nets} also approximate the rounding function with  a stack of sigmoid functions to address the \textit{gradient mismatch} problem. They parameterize the sigmoid with a temperature, which increases during training to better approximate the step function.

As we show in the appendix, these multiplicative methods can affect the magnitude of the oscillations but cannot stop them from happening. In figure \ref{fig:oscillation_single_weight}, we visualize the effect of EWGS and DSQ gradients on the 1D regression problem, described in section \ref{sec:oscillations}. It is evident, that neither of these methods overcome oscillations.
\vspace{-.1cm}
\paragraph{Additive} Various regularizations have been proposed in literature to make the weight distribution more quantization-friendly.
\citet{qat_tinyml} apply an $L^2$ regularization on the difference between quantized and floating-point weights, while \citet{binreg} apply bin regularization to enforce the weights to cluster at the quantization grid points. Their bin regularization method is similar to our proposed \textit{oscillation dampening} method described later in section \ref{sec:oscillations_dampening}. However, their motivation is to form sharper weight distribution around the bin centre that reduce the quantization error rather than preventing oscillations. In contrast to the multiplicative methods, these additive methods add a term to the gradient and can thus dampen oscillations by pulling the latent weights towards the grid points.


\citet{profit} recognize that the STE can cause activation instability during training of low-bit quantized MobileNets and propose progressively freezing whole layers during training. \citet{metaquant} approximate the gradient using a small neural network, whose parameters are learnt using meta-learning but is limited to binary weights.

\section{Overcoming oscillations in QAT}\label{sec:method}
Now that we have established that oscillations can have a negative effect during optimization, especially for low-bit quantization, we focus on how to overcome them.
First we introduce a metric for quantifying oscillations and then we propose two novel techniques aimed at preventing oscillations at their source during quantization-aware training.  

\subsection{Quantifying Oscillations}
\label{sec:quantifying_oscillations}
Before we can address the oscillations, we need a way of detecting and measuring them during training. We propose calculating the frequency of oscillations over time using an exponential moving average (EMA). We can then define a minimum frequency as a threshold for oscillating weights. 
For an oscillation to occur in iteration $t$ it needs two satisfy two conditions: 
\begin{enumerate}
    \item The integer value of the weight needs to change, thus $\vec{w}^t_{\INT} \neq \vec{w}^{t-1}_{\INT} $ where $\vec{w}^t_{\INT} =  \text{clip} \left(\round*{\frac{\vec{w}}{\ct{s}}}, \ct{n}, \ct{p}\right)$ is the integer value of the quantized weight.
    \item The direction of the change in the integer domain needs to be the opposite than that of its previous change, thus $o^t = \text{sign}(\Delta^t_{\INT}) \neq \text{sign}(\Delta^{\tau}_{\INT})$ where $\tau$ is the iteration of the last change in integer domain and $\Delta^t_{\INT} = \vec{w}^t_{\INT} - \vec{w}^{t-1}_{\INT}$, the direction of the change.
\end{enumerate}
We then track the frequency of oscillations over time using an exponential moving average (EMA):
\begin{equation}
    f^t = m \cdot o^t + (1-m) \cdot f^{t-1}.
    \label{eq:track_oscillations}
\end{equation}

\subsection{Oscillation dampening}
\label{sec:oscillations_dampening}
When weights oscillate, they always move around the decision threshold between two quantization bins. This means that oscillating weights are always close to the edge of the quantization bin. In order to dampen the oscillatory behavior, we employ a regularization term that encourages latent weights to be close to the center of the bin rather than its edge. We define our dampening loss similar to weight decay as:
\begin{equation}
    \label{eq:loss_dampen}
    \tlossb_{\text{dampen}} = \norm*{\vecq{w} - \clip{\vec{w}}{ s \ct{n}}{s \ct{p}}}^2_F,
\end{equation}
where $s$, $n$ and $p$ are the quantization parameters as defined in equation \eqref{eq:quantization_formulation}
and $\vecq{w}$ are the bin centers. Note, as the bin centers are our optimization target, no gradients propagate back through this term.
Our final training objective is now: $\tlossb = \tlossb_{\text{task}} + \lambda\tlossb_{\text{dampen}}$. We choose to apply our bin regularization in the latent weight domain such that the resulting gradient
\begin{equation}
    \frac{\partial\tlossb_{\text{dampen}}}{\partial \vec{w}} = 2 \l(\vec{w} - \vecq{w} \r ) \cdot  \mathbf{1}_{\ct{s}n \leq \vec{w} \leq \ct{s}p}
\end{equation}
is independent of the scale $s$ and, therefore, indirectly independent of the bit-width\footnote{A similar loss defined in integer domain ($\norm*{\vec{w}_{\INT} - \vec{w}/s}^2_F$) would lead to a scaling of gradient with $1/s^2$ and make the regularization on $\vec{w}$ dependent on the scale and bit-width.}.
We further clip our latent weights to the range of the quantization grid, such that only weights that do not get clipped during quantization receive a regularization effect. This is important to avoid any harmful interactions with quantization scale gradient in LSQ-based range learning \citep{lsq}.  A drawback of such a regularization is that it does not only affect weights that oscillate but can also hinder the movement of weights that are not in an oscillating state.



\subsection{Iterative freezing of oscillating weights}
\begin{algorithm}[t]
	\caption{QAT with iterative weight freezing}
	\label{alg:qat_freezing}
	\begin{algorithmic}[1]
	    \STATE Init: $f^0 \gets \vec{0}, b \gets \vec{0}, \Delta^{\tau} \gets \vec{0}, \vec{w}^0_{\text{EMA(int)}} \gets \vec{w}_\INT^0$
	    \FOR{t = 1,\dots, T}
	        \STATE Calculate gradient $g^t = \frac{\partial\vec{\tlossb}}{\partial \vec{w}}$
	        \STATE Optimizer update for weights $\vec{w}^t[\neg b]$ using $g^t$
	        \STATE $\vec{w}^t_{\INT} \gets  \text{clip} \left(\round*{\frac{\vec{w}^t}{\ct{s}}}, \ct{n}, \ct{p}\right)$
	        \STATE $\Delta^t_{\INT} \gets \vec{w}^t_{\INT} - \vec{w}^{t-1}_{\INT}$
	        \STATE $o^t \gets (\text{sign}(\Delta^t_{\INT}) \neq \text{sign}(\Delta^{\tau}_{\INT})) \odot (\Delta^t_{\INT} \neq 0) $
	        \STATE $f^t \gets m \cdot o^t + (1-m) \cdot f^{t-1}$
	        \FOR{i = 1,\dots, N}
	            \IF{$f^t_i > f_{\text{th}}$}
	                \STATE $b_i \gets \text{True}$ 
	                \STATE $\vec{w}_i^{t} \gets \ct{s} \cdot \round*{\vec{w}^{t-1}_{\text{EMA(int)}_i}}$  
	            \ENDIF
	        \ENDFOR
            \STATE $\vec{w}^t_{\text{EMA(int)}} \gets m \cdot \vec{w}^{t-1} + (1-m) \cdot \vec{w}^{t-1}_{\text{EMA(int)}}$\label{lst:weigth_ema}
	        \STATE $\Delta^{\tau}_{\INT}[o^t] \gets \Delta^t_{\INT}[o^t]$
	            
		\ENDFOR
	
	\end{algorithmic} 
\end{algorithm}

\label{sec:freezing_method}
We propose another, more targeted, approach to prevent weights from oscillating by freezing them during training.
In this method, we track the oscillations frequency per weight during training, as described in equation \eqref{eq:track_oscillations}. If the oscillation frequency of any weight exceeds a threshold $f_{\text{th}}$, that weight gets frozen until the end of training. We apply the freezing in the  integer domain, such that potential change in the scale $s$ during optimization does not lead to a different rounding. 

When a weight oscillates, it does not necessarily spend an equal amount of time at both oscillating states. As we show in the toy example in section \ref{sec:oscillations}, the likelihood of a weight being in each state linearly depends on distance of that quantized state from the optimal value. As a result, the expectation of all quantized values over time will correspond to the optimal value. Once the frequency of a weight exceeds the threshold, it could be in either of the two quantized states. To freeze the weight to its more frequent state we keep a record of the previous integer values using an exponential moving average (EMA). We then assign the most frequent integer state to the frozen weight by rounding the EMA.


We summarize our proposed iterative weight freezing in algorithm \ref{alg:qat_freezing}. Note, this algorithm can be used in combination with any gradient-based optimizer and is not limited to a particular quantization formulation or gradient estimator. The idea of freezing weights on an iteration level is closely related to iterative pruning \citep{zhu2017prune}, where small weights are iteratively pruned (frozen to zero).


\section{Experiments}\label{sec:experiments}
In this section, we evaluate the effectiveness of our proposed methods for overcoming oscillations and compare them against other QAT methods on ImageNet \citep{imagenet}. We focus on low-bit quantization, 4 and 3 bits, of efficient networks with depth-wise separable convolutions, where the effect of weight oscillation is most pronounced. We start with our ablation studies, where we only quantize the weights in a network and, finally, present results for low-bit weight and activation quantization.

\subsection{Experimental setup}

\label{sec:experimentalconv.3.1_setup}


\paragraph{Quantization}
We follow the example of existing QAT literature and apply LSQ-type \citep{lsq} weight and activation quantization: we quantize all weights to a low bit-width while keeping the weights of the first and last layer to 8-bits, and quantize the input to all layers expect for the normalizing layers. We use per-tensor quantization \cite{krishnamoorthi} and learn the quantization scaling factor.

\paragraph{Optimization}
In all cases, we start from a pre-trained full-precision network and instantiate the weight and activation quantization parameters using MSE range estimation \citep{whitepaper}. We use SGD with a momentum of 0.9 and train using a cosine annealing learning-rate decay. We train for 20 epochs with only weight quantization for the ablation studies. For weight and activation quantization in section \ref{sec:comparison_to_qat_methods}, we train all models for 90 epochs. Depending on the network and quantization bit-width we train with a learning rate of either 0.01 or 0.0033.


\subsection{Ablation studies}
\label{sec:ablation_studies}
\paragraph{Oscillation dampening}
In table \ref{tbl:ocillation_dampening_ablation} we study how the strength of the dampening loss affects the network's final accuracy and the proportion of oscillating weights at the end of training. In the first three rows, we observe that as we increase the coefficient $\lambda$, the proportion of oscillating weights decreases and the gap between pre and post-BN re-estimation accuracy closes. However, too much dampening harms the final accuracy, suggesting that excessive regularization inhibits beneficial movement of weights between quantization levels.

Our solution to this problem is to gradually increase the regularization strength during training. This allows the latent weights to move more freely in the first stages of training, while reducing harmful oscillations close to convergence by applying a stronger regularization. We find that a cosine annealing schedule for $\lambda$ works well in practice. \citet{binreg} also notice that such a regularization is harmful at the early stages of training but instead adopt a two-stage optimization process. We see that such a strategy can significantly dampen the oscillations while not harming the accuracy. The best damping configuration improves almost 1\% over the post-BN re-estimation baseline and more than 5\% over the pre-BN re-estimation baseline.

\begin{figure}[t]
\centering
\includegraphics[width=1.\linewidth]{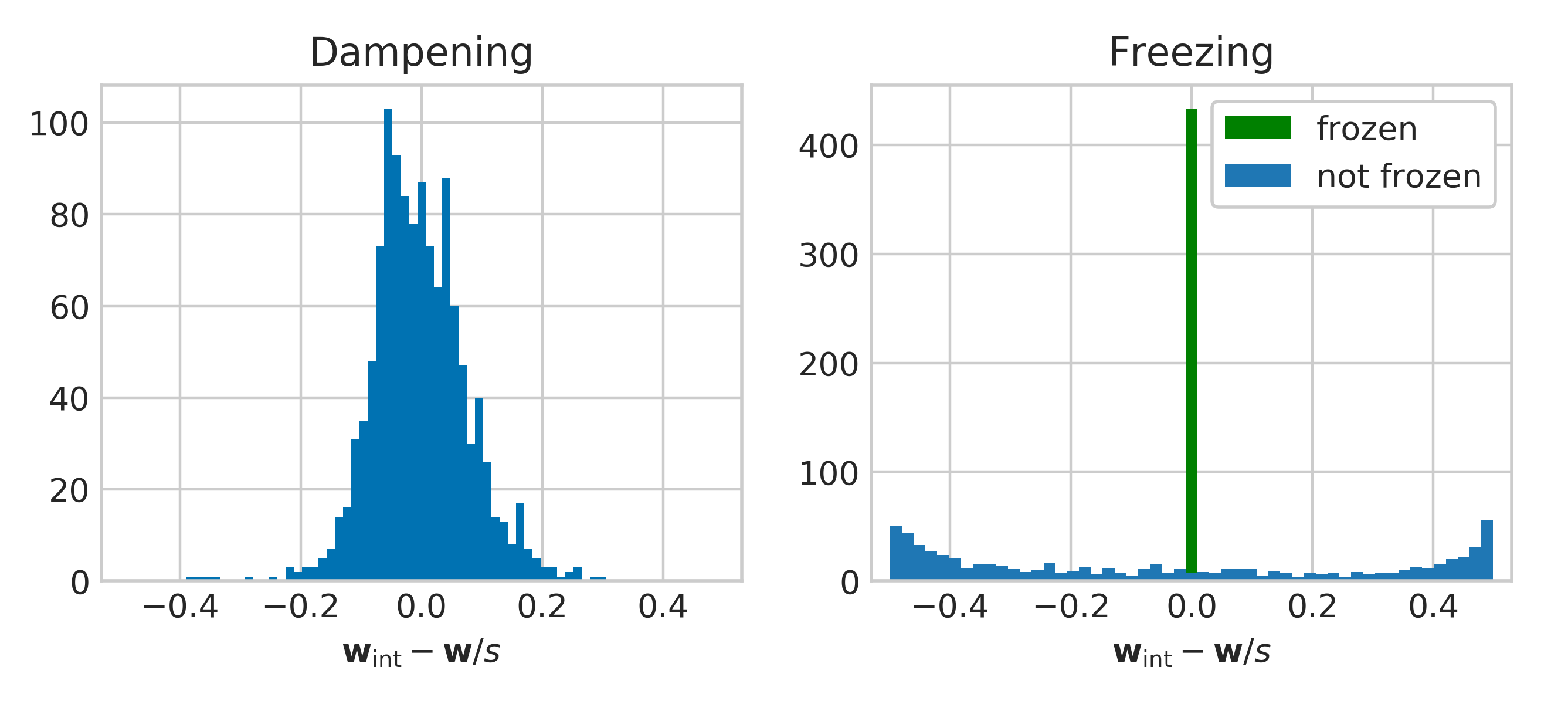} 
\vspace{-0.8cm}
\caption{ Histogram of the distance of latent weights of layer conv.3.1 of \mnv{2} from their closest quantization grid-point after training with oscillation dampening (left) or iterative weight freezing (right).} 
\vspace{-0.1cm}
\label{fig:bin_distribution_freezing_damp}
\end{figure}

In figure \ref{fig:bin_distribution_freezing_damp} (left), we also visualize the effect of dampening on the latent weight distribution of the same depth-wise separable layer as in figure \ref{fig:mv2_bin_distribution}. As intended, the latent weights are now clustered around the quantization bin centres with hardly any weights at the decision boundary. 

\begin{table}[t]
    \centering
    \begin{tabular}{ l | c  c  c }
    \toprule
  Regulatization & pre-BN &  post-BN & Osc.(\%) \\
        \midrule
Baseline  & 64.97\textsuperscript{1.23} & 69.50\textsuperscript{0.04} & 4.93\\
\midrule
$\lambda=10^{-4}$   & 65.97\textsuperscript{1.52}       &  69.65\textsuperscript{0.08} &  2.18 \\
$\lambda=10^{-3}$   & {66.99}\textsuperscript{1.41}     & \textbf{69.96}\textsuperscript{0.12} & 0.21 \\
$\lambda=10^{-2}$	& \textbf{68.04}\textsuperscript{1.04}  & 68.57\textsuperscript{0.07} & 0.01 \\
\midrule
$\lambda=\cos(0, 10^{-4})$  & 64.47\textsuperscript{1.59} & 69.61\textsuperscript{0.07} & 2.64 \\ 
$\lambda=\cos(0, 10^{-3})$  & {68.79}\textsuperscript{1.31} & \textbf{70.37}\textsuperscript{0.06} & 1.63 \\ 
$\lambda=\cos(0, 10^{-2})$  & \textbf{70.18}\textsuperscript{0.18} & 70.26\textsuperscript{0.08} & 1.11 \\ 
\bottomrule
\end{tabular}
 \caption{Effect of oscillation dampening strength and schedule in the final accuracy of \mnv{2} with 3-bit weight quantization. \textit{post-BN}: accuracy after BN re-estimation, \textit{pre-BN}: accuracy before BN re-estimation. \textit{Osc.}: percentage of oscillating weights at the end of training ($f>0.005$). We report the average of 3 seeds and the STD in the superscript.}
 \label{tbl:ocillation_dampening_ablation}
\end{table}

\begin{table}[t]
    \centering
    \begin{tabular}{ l | c  c c }
    \toprule
  Method &   pre-BN & post-BN & Osc.(\%)\\
        \midrule
Baseline & 64.97\textsuperscript{1.23} & 69.50\textsuperscript{0.04} &  4.93 \\
\midrule
$f_{\text{th}}$= 0.02 & 68.13\textsuperscript{2.14}  & 69.96\textsuperscript{0.04} &   2.93 \\
$f_{\text{th}}$= 0.015 & \textbf{69.79}\textsuperscript{0.07} & \textbf{70.13}\textsuperscript{0.05} &  1.23\\
$f_{\text{th}}$= 0.01  &  69.12\textsuperscript{0.53} & 69.18\textsuperscript{0.47}  &   0.06 \\
\midrule
$f_{\text{th}}$= cos(0.04,0.015)  &  69.51\textsuperscript{0.15}  & 69.96\textsuperscript{0.03} &  2.33 \\
$f_{\text{th}}$= cos(0.04,0.01)  &  \textbf{69.97}\textsuperscript{0.06}  & \textbf{70.33}\textsuperscript{0.07} &  0.04 \\
\bottomrule
\end{tabular}
 \caption{Effect of oscillation frequency threshold ($f_{\text{th}}$) on training of \mnv{2} with 3-bit weight quantization. \textit{post-BN}: accuracy after BN re-estimation, \textit{pre-BN}: accuracy before BN re-estimation, \textit{Osc.}: percentage of oscillating weights at the end of training ($f>0.005$). We report the average of 3 seeds and the STD in the superscript.}
 \label{tbl:weight_freezing_ablation}
\end{table}

\paragraph{Iterative weight freezing}
In table \ref{tbl:weight_freezing_ablation}, we demonstrate the effectiveness of our iterative weight freezing algorithm for various freezing thresholds. Using a constant threshold throughout training, we see that the amount of remaining oscillations significantly reduces with a decreasing threshold and only some low frequency oscillations remain in the network. We also see that the pre-BN re-estimation accuracy is  closer to the the post-BN re-estimation accuracy, as one would expect when there are fewer oscillations at the end of training. 

\begin{table}[t]
    \centering
    \begin{tabular}{ l | c |  c }
    \toprule

  Method & W/A &    Val. Acc. (\%)  \\
        \midrule
Full-precision & 32/32 & 71.7 \\
\midrule
LSQ* \citep{lsq}         & 4/4 & 69.5 (-2.3) \\ 
PACT \cite{pact2018}    & 4/4 & 61.4  (-10.3)\\ 
DSQ \citep{DSQ}         & 4/4 & 64.8 (-6.9) \\
EWGS \citep{EWGS}       & 4/4 & 70.3 (-1.6) \\ 
LSQ + BR \citep{binreg} & 4/4 & 70.4 (-1.4)\\ 
LSQ + Dampen (ours)     & 4/4 & \textbf{70.5} (-1.2) \\
LSQ + Freeze (ours)     & 4/4 & \textbf{70.6} (-1.1) \\
\midrule
LSQ* \citep{lsq}         & 3/3 & 65.3 (-6.5)  \\
LSQ + BR \citep{binreg} & 3/3 & 67.4 (-4.4) \\
LSQ + Dampen (ours)     & 3/3 & \textbf{67.8} (-3.9)  \\
LSQ + Freeze (ours)     & 3/3 & \textbf{67.6} (-4.1) \\
\bottomrule
\end{tabular}
 \caption{Comparison of oscillation dampening and freezing with other QAT methods for \mnv{2} on ImageNet classification. We report validation accuracy (\%) and the difference to the corresponding full precision baseline. Note, both \citet{binreg} and \citet{EWGS} use a slightly higher baseline model. *results from \citet{binreg}.}
 \vspace{-0.1cm}
 \label{tbl:final_results_mnv2}
\end{table}

\begin{table}[t]
    \centering
    \begin{tabular}{ l | c |  c }
    \toprule

  Method & W/A &    Val. Acc. (\%)  \\
        \midrule
Full-precision & 32/32 & 65.1 \\
\midrule
LSQ* \citep{lsq}         & 4/4 & 61.0  \\
LSQ + BR \citep{binreg} & 4/4 & 61.5 \\
LSQ + Dampen (ours)     & 4/4 & \textbf{63.7}  \\ 
LSQ + Freeze (ours)     & 4/4 &  \textbf{63.6} \\
\midrule
LSQ* \citep{lsq}         & 3/3 &  52.0 \\
LSQ + BR \citep{binreg} & 3/3 &   56.0 \\
LSQ + Dampen (ours)     & 3/3 &   \textbf{59.0}  \\
LSQ + Freeze (ours)     & 3/3 &   \textbf{58.9} \\
\bottomrule
\end{tabular}
 \caption{Comparison of oscillation dampening and freezing with other QAT methods for \mnv{3}-Small on ImageNet classification. We report validation accuracy (\%). *results from \citet{binreg}.}
 \label{tbl:final_results_mnv3}
\end{table}

\begin{table}[t]
    \centering
    \begin{tabular}{ l | c |  c }
    \toprule

  Method & W/A &    Val. Acc. (\%)  \\
        \midrule
Full-precision          & 32/32 & 75.4 \\
\midrule
LSQ* \citep{lsq}         & 4/4 &  72.3 \\
LSQ + Dampen (ours)     & 4/4 &  \textbf{73.5} \\
LSQ + Freeze (ours)     & 4/4 &  \textbf{73.5} \\
\midrule
LSQ* \citep{lsq}         & 3/3 &  69.7 \\
LSQ + Dampen (ours)     & 3/3 &  \textbf{71.1}  \\
LSQ + Freeze (ours)     & 3/3 &  \textbf{71.0} \\
\bottomrule
\end{tabular}
 \caption{Comparison of oscillation dampening and freezing for EfficientNet-lite on ImageNet classification. We report validation accuracy (\%). *our re-implementation.}
 \label{tbl:final_results_en_lite}
\end{table}



However, we note that if the oscillation threshold becomes too low, too many weights get frozen in the early stages of training reducing the final accuracy. To address this, we apply an annealing schedule to the freezing threshold similar to the one used in dampening. This allows us to use a stronger freezing threshold and freeze almost all oscillations towards the end of training, when they are most disruptive.  The best freezing threshold improves almost 1\% over the post BN re-estimation baseline and more than 5\% over the pre BN re-estimation baseline. Its accuracy is on par with the oscillation dampening while having significantly less remaining oscillations (0.04\% vs 1.11\%).

In figure \ref{fig:bin_distribution_freezing_damp} (right), we see how iterative weight freezing changes the latent weight distribution of layer conv.3.1 of \mnv{2}. The majority of the latent weights are now frozen at the bin centres removing the peaks observed at the decision boundary in figure \ref{fig:mv2_bin_distribution}.



\subsection{Comparison to other QAT methods}
\label{sec:comparison_to_qat_methods}
We compare our methods for overcoming oscillation  to other QAT alternatives and demonstrate their effectiveness in low-bit quantization of popular efficient neural networks. To compare to existing methods in the literature we quantize both weights and activations. 
In table \ref{tbl:final_results_mnv2}, we show the results for \mnv{2} and demonstrate that both our algorithms outperform all competing QAT techniques in literature for 3 and 4-bit quantization. We also achieve state-of-the-art results for \mnv{3}-Small and \efnet{lite} in tables \ref{tbl:final_results_mnv3} and \ref{tbl:final_results_en_lite}, respectively. In all cases, our oscillation prevention methods significantly improve ($>1\%$) over the commonly used LSQ baseline, showing the general applicability of our methods to other efficient networks.

We note that oscillation dampening leads to an increase of approximately 33\% in training time compared to the LSQ baseline. A similar trend was also observed by \cite{binreg}. On the other hand, iterative weight freezing has a negligible computational overhead while achieving similar performance.

\section{Conclusion}\label{sec:conclusion}
In this work, we showed that weight oscillations induced by the straight-through estimator and its variations proposed in literature can adversely affect the performance of quantized networks when trained with quantization-aware training. The effect of weight oscillations is more pronounced in low-bit quantization of light-weight networks, such as \mnv{2} that include depth-wise layers with only a few weights per output channel. 
We demonstrate that these oscillations not only corrupt the running batch-normalization  statistics used during inference but also harm the optimization process itself, leading to sub-optimal solutions. We propose two methods to tackle this problem: \textit{oscillations dampening} and \textit{iterative weight freezing}, and  show that both methods lead to state-of-the-art accuracy on ImageNet classification for a range of efficient models quantized to low-bits. 
\FloatBarrier


\bibliography{dirty}

\begin{thebibliography}{38}
\providecommand{\natexlab}[1]{#1}
\providecommand{\url}[1]{\texttt{#1}}
\expandafter\ifx\csname urlstyle\endcsname\relax
  \providecommand{\doi}[1]{doi: #1}\else
  \providecommand{\doi}{doi: \begingroup \urlstyle{rm}\Url}\fi

\bibitem[Banner et~al.(2019)Banner, Nahshan, and Soudry]{bannerposttraining}
Banner, R., Nahshan, Y., and Soudry, D.
\newblock Post training 4-bit quantization of convolutional networks for
  rapid-deployment.
\newblock In \emph{Advances in Neural Information Processing Systems
  (NeuRIPS)}, 2019.

\bibitem[Bengio et~al.(2013)Bengio, L{\'e}onard, and
  Courville]{bengio2013estimating}
Bengio, Y., L{\'e}onard, N., and Courville, A.
\newblock Estimating or propagating gradients through stochastic neurons for
  conditional computation.
\newblock \emph{arXiv preprint arXiv:1308.3432}, 2013.

\bibitem[Bhalgat et~al.(2020)Bhalgat, Lee, Nagel, Blankevoort, and Kwak]{lsq+}
Bhalgat, Y., Lee, J., Nagel, M., Blankevoort, T., and Kwak, N.
\newblock Lsq+: Improving low-bit quantization through learnable offsets and
  better initialization.
\newblock In \emph{Proceedings of the IEEE/CVF Conference on Computer Vision
  and Pattern Recognition (CVPR) Workshops}, 2020.

\bibitem[Cai et~al.(2017)Cai, He, Sun, and Vasconcelos]{guassian_quant}
Cai, Z., He, X., Sun, J., and Vasconcelos, N.
\newblock Deep learning with low precision by half-wave gaussian quantization.
\newblock In \emph{Conference on Computer Vision and Pattern Recognition
  (CVPR)}, 2017.

\bibitem[Chai(2021)]{qat_tinyml}
Chai, S.~M.
\newblock Quantization-guided training for compact tiny{\{}ml{\}} models.
\newblock In \emph{Research Symposium on Tiny Machine Learning}, 2021.

\bibitem[Chen et~al.(2019)Chen, Wang, and Pan]{metaquant}
Chen, S., Wang, W., and Pan, S.~J.
\newblock Metaquant: Learning to quantize by learning to penetrate
  non-differentiable quantization.
\newblock In \emph{Neural Information Processing Systems (NeuRIPS)}, 2019.

\bibitem[Chmiel et~al.(2022)Chmiel, Banner, Hoffer, Yaacov, and
  Soudry]{chmiel2022luq}
Chmiel, B., Banner, R., Hoffer, E., Yaacov, H.~B., and Soudry, D.
\newblock Logarithmic unbiased quantization: Practical 4-bit training in deep
  learning.
\newblock \emph{arXiv preprint arXiv:2112.10769}, 2022.

\bibitem[Choi et~al.(2018)Choi, Wang, Venkataramani, Chuang, Srinivasan, and
  Gopalakrishnan]{pact2018}
Choi, J., Wang, Z., Venkataramani, S., Chuang, P.~I., Srinivasan, V., and
  Gopalakrishnan, K.
\newblock {PACT:} parameterized clipping activation for quantized neural
  networks.
\newblock \emph{arXiv preprint arxiv:805.06085}, 2018.

\bibitem[Défossez et~al.(2021)Défossez, Adi, and Synnaeve]{diffq}
Défossez, A., Adi, Y., and Synnaeve, G.
\newblock Differentiable model compression via pseudo quantization noise.
\newblock \emph{arXiv preprint arXiv:2104.09987}, 2021.

\bibitem[Esser et~al.(2020)Esser, McKinstry, Bablani, Appuswamy, and
  Modha]{lsq}
Esser, S.~K., McKinstry, J.~L., Bablani, D., Appuswamy, R., and Modha, D.~S.
\newblock Learned step size quantization.
\newblock In \emph{International Conference on Learning Representations
  (ICLR)}, 2020.

\bibitem[Fan et~al.(2021)Fan, Stock, Graham, Grave, Gribonval, Jegou, and
  Joulin]{fan2021extreme}
Fan, A., Stock, P., Graham, B., Grave, E., Gribonval, R., Jegou, H., and
  Joulin, A.
\newblock Training with quantization noise for extreme model compression.
\newblock In \emph{International Conference on Learning Repersentations
  (ICLR)}, 2021.

\bibitem[Gong et~al.(2019)Gong, Liu, Jiang, Li, Hu, Lin, Yu, and Yan]{DSQ}
Gong, R., Liu, X., Jiang, S., Li, T., Hu, P., Lin, J., Yu, F., and Yan, J.
\newblock Differentiable soft quantization: Bridging full-precision and low-bit
  neural networks.
\newblock \emph{International Conference on Computer Vision (ICCV)}, 2019.

\bibitem[Gupta et~al.(2015)Gupta, Agrawal, Gopalakrishnan, and
  Narayanan]{Gupta2015}
Gupta, S., Agrawal, A., Gopalakrishnan, K., and Narayanan, P.
\newblock Deep learning with limited numerical precision.
\newblock In \emph{International Conference on Machine Learning, (ICML)}, 2015.

\bibitem[Han et~al.(2021)Han, Li, Liu, Tian, and Shan]{binreg}
Han, T., Li, D., Liu, J., Tian, L., and Shan, Y.
\newblock Improving low-precision network quantization via bin regularization.
\newblock In \emph{International Conference on Computer Vision (ICCV)}, 2021.

\bibitem[Helwegen et~al.(2019)Helwegen, Widdicombe, Geiger, Liu, Cheng, and
  Nusselder]{helwegen2019latent}
Helwegen, K., Widdicombe, J., Geiger, L., Liu, Z., Cheng, K.-T., and Nusselder,
  R.
\newblock Latent weights do not exist: Rethinking binarized neural network
  optimization.
\newblock In \emph{Advances in Neural Information Processing Systems
  (NeuRIPS)}, 2019.

\bibitem[Hinton(2012)]{hinton2012ste}
Hinton, G.
\newblock Neural networks for machine learning, lectures 15b.
\newblock 2012.

\bibitem[{Horowitz}(2014)]{horowitz}
{Horowitz}, M.
\newblock 1.1 computing's energy problem (and what we can do about it).
\newblock In \emph{2014 IEEE International Solid-State Circuits Conference
  Digest of Technical Papers (ISSCC)}, pp.\  10--14, 2014.

\bibitem[Ioffe \& Szegedy(2015)Ioffe and Szegedy]{batchnorm}
Ioffe, S. and Szegedy, C.
\newblock Batch normalization: Accelerating deep network training by reducing
  internal covariate shift.
\newblock In \emph{International Conference on Machine Learning (ICML)}, 2015.

\bibitem[J.~Lee(2021)]{EWGS}
J.~Lee, D.~Kim, B.~H.
\newblock Network quantization with element-wise gradient scaling.
\newblock In \emph{Conference on Computer Vision and Pattern Recognition
  (CVPR)}, 2021.

\bibitem[Jain et~al.(2019)Jain, Gural, Wu, and Dick]{tqt}
Jain, S.~R., Gural, A., Wu, M., and Dick, C.
\newblock Trained uniform quantization for accurate and efficient neural
  network inference on fixed-point hardware.
\newblock \emph{arxiv preprint arxiv:1903.08066}, 2019.

\bibitem[Kim et~al.(2020)Kim, Yoo, and Kwak]{PBGS}
Kim, J., Yoo, K., and Kwak, N.
\newblock Position-based scaled gradient for model quantization and pruning.
\newblock In \emph{Advances in Neural Information Processing Systems
  (NeuRIPS)}, 2020.

\bibitem[Kirkpatrick et~al.(1983)Kirkpatrick, Gelatt, and
  Vecchi]{Kirkpatrick1983SimulatedAnnealing}
Kirkpatrick, S., Gelatt, C., and Vecchi, M.
\newblock Optimization by simulated annealing.
\newblock \emph{Science (New York, N.Y.)}, 220:\penalty0 671--80, 06 1983.

\bibitem[{Krishnamoorthi}(2018)]{krishnamoorthi}
{Krishnamoorthi}, R.
\newblock {Quantizing deep convolutional networks for efficient inference: A
  whitepaper}.
\newblock \emph{arXiv preprint arXiv:1806.08342}, 2018.

\bibitem[Louizos et~al.(2019)Louizos, Reisser, Blankevoort, Gavves, and
  Welling]{louizos2018relaxed}
Louizos, C., Reisser, M., Blankevoort, T., Gavves, E., and Welling, M.
\newblock Relaxed quantization for discretized neural networks.
\newblock In \emph{International Conference on Learning Representations
  (ICLR)}, 2019.

\bibitem[Nagel et~al.(2019)Nagel, van Baalen, Blankevoort, and Welling]{dfq}
Nagel, M., van Baalen, M., Blankevoort, T., and Welling, M.
\newblock Data-free quantization through weight equalization and bias
  correction.
\newblock In \emph{International Conference on Computer Vision (ICCV)}, 2019.

\bibitem[Nagel et~al.(2020)Nagel, Amjad, Van~Baalen, Louizos, and
  Blankevoort]{adaround}
Nagel, M., Amjad, R.~A., Van~Baalen, M., Louizos, C., and Blankevoort, T.
\newblock Up or down? {A}daptive rounding for post-training quantization.
\newblock In \emph{International Conference on Machine Learning (ICML)}, 2020.

\bibitem[Nagel et~al.(2021)Nagel, Fournarakis, Amjad, Bondarenko, van Baalen,
  and Blankevoort]{whitepaper}
Nagel, M., Fournarakis, M., Amjad, R.~A., Bondarenko, Y., van Baalen, M., and
  Blankevoort, T.
\newblock A white paper on neural network quantization.
\newblock \emph{arXiv preprint arXiv:2106.08295}, 2021.

\bibitem[Nahshan et~al.(2020)Nahshan, Chmiel, Baskin, Zheltonozhskii, Banner,
  Bronstein, and Mendelson]{nahshan2020lapq}
Nahshan, Y., Chmiel, B., Baskin, C., Zheltonozhskii, E., Banner, R., Bronstein,
  A.~M., and Mendelson, A.
\newblock Loss aware post-training quantization.
\newblock \emph{arXiv preprint arXiv:1911.07190}, 2020.

\bibitem[Nguyen et~al.(2020)Nguyen, Alexandridis, and
  Mouchtaris]{qat_abs_cosine}
Nguyen, H.~D., Alexandridis, A., and Mouchtaris, A.
\newblock Quantization aware training with absolute-cosine regularization for
  automatic speech recognition.
\newblock In \emph{Interspeech}, 2020.

\bibitem[Park \& Yoo(2020)Park and Yoo]{profit}
Park, E. and Yoo, S.
\newblock {PROFIT:} {A} novel training method for sub-4-bit mobilenet models.
\newblock In \emph{European Conference on Computer Vision (ECCV)}, 2020.

\bibitem[Pervez et~al.(2020)Pervez, Cohen, and Gavves]{pervez2020lowbl}
Pervez, A., Cohen, T., and Gavves, E.
\newblock Low bias low variance gradient estimates for hierarchical boolean
  stochastic networks.
\newblock In \emph{International Conference on Machine Learning (ICML)}, 2020.

\bibitem[Peters \& Welling(2018)Peters and Welling]{peters2018probabilistic}
Peters, J. W.~T. and Welling, M.
\newblock Probabilistic binary neural networks.
\newblock \emph{arXiv preprint arXiv:1809.03368}, 2018.

\bibitem[Russakovsky et~al.(2015)Russakovsky, Deng, Su, Krause, Satheesh, Ma,
  Huang, Karpathy, Khosla, Bernstein, Berg, and Fei-Fei]{imagenet}
Russakovsky, O., Deng, J., Su, H., Krause, J., Satheesh, S., Ma, S., Huang, Z.,
  Karpathy, A., Khosla, A., Bernstein, M., Berg, A.~C., and Fei-Fei, L.
\newblock {ImageNet Large Scale Visual Recognition Challenge}.
\newblock \emph{International Journal of Computer Vision (IJCV)}, 2015.

\bibitem[Uhlich et~al.(2020)Uhlich, Mauch, Cardinaux, Yoshiyama, Garcia,
  Tiedemann, Kemp, and Nakamura]{differentiablequantization}
Uhlich, S., Mauch, L., Cardinaux, F., Yoshiyama, K., Garcia, J.~A., Tiedemann,
  S., Kemp, T., and Nakamura, A.
\newblock Mixed precision dnns: All you need is a good parametrization.
\newblock In \emph{International Conference on Learning Representations
  (ICLR)}, 2020.

\bibitem[Yang et~al.(2019)Yang, Shen, Xing, Tian, Li, Deng, Huang, and
  Hua]{quantization_nets}
Yang, J., Shen, X., Xing, J., Tian, X., Li, H., Deng, B., Huang, J., and Hua,
  X.-s.
\newblock Quantization networks.
\newblock In \emph{Conference on Computer Vision and Pattern Recognition
  (CVPR)}, 2019.

\bibitem[Yin et~al.(2019)Yin, Lyu, Zhang, Osher, Qi, and
  Xin]{understanding_STE}
Yin, P., Lyu, J., Zhang, S., Osher, S.~J., Qi, Y., and Xin, J.
\newblock Understanding straight-through estimator in training activation
  quantized neural nets.
\newblock In \emph{International Conference on Learning Representations(
  ICLR)}, 2019.

\bibitem[Zhou et~al.(2016)Zhou, Ni, Zhou, Wen, Wu, and Zou]{dorefa}
Zhou, S., Ni, Z., Zhou, X., Wen, H., Wu, Y., and Zou, Y.
\newblock Dorefa-net: Training low bitwidth convolutional neural networks with
  low bitwidth gradients.
\newblock \emph{arXiv preprint arXiv:1606.06160}, 2016.

\bibitem[Zhu \& Gupta(2018)Zhu and Gupta]{zhu2017prune}
Zhu, M. and Gupta, S.
\newblock To prune, or not to prune: exploring the efficacy of pruning for
  model compression.
\newblock In \emph{International Conference on Learning Representations,
  {ICLR}, Workshop Track Proceedings}, 2018.

\end{thebibliography}
\bibliographystyle{icml2022}

\newpage
\appendix
\onecolumn
\section{Appendix}
\label{sec:appendix}
\subsection{Gradients for 1D regression problem}
\label{app:toy_example_gradients}
In this section, we analytically derive the gradient descent (GD) updates for the latent weight the 1D toy regression example of section \ref{sec:oscillations}.
We compare the gradient from the vanilla STE with other methods from the literature (cf. section \ref{sec:backgroundrelated}) and our proposed oscillation dampening solution (cf. section \ref{sec:oscillations_dampening}). We define $\bar{w}$ as the boundary between the two quantization levels closest to the optimal value $w_*$, such that $\bar{w}= (\up{w}+\down{w})/2$, where $\up{w}=q(w_*+\frac{s}{2})$  and $\down{w}=q(w_*-\frac{s}{2})$.

The gradient of the mean-squared loss from equation \eqref{eq:toy_regression_optim} w.r.t. to the weight is given by: 
\begin{align} 
    \frac{\partial \tlossb(w)}{\partial w}& =\sigma^2 \left(q(w) - w_* \right)  \frac{\partial q(w)}{\partial w }, \\
    &= \begin{cases}
    \sigma^2 \left(\up{w} - w_*\right)\frac{\partial q(w)}{\partial w } , & \text{if }w \geq \bar{w} \\
    \sigma^2 \left(\down{w} - w_* \right) \frac{\partial q(w)}{\partial w },& \text{if } w < \bar{w}\\
    \end{cases} 
    \label{eq:regression_grad_quant}
\end{align}

Assuming $\sigma=1$, a learning rate of $\eta$, and that no clipping takes place we write the GD update rules for subset of methods discussed in section \ref{sec:backgroundrelated}. For the PSG \citep{PBGS}, the update rule includes an arbitrarily small constant, $\epsilon>0$, while EWGS  \citep{EWGS} introduce a $\delta \geq 0$ scaling factor. 

\begin{center}
\begin{tabular}{ l | l } 
STE &  
$w_{t+1} = 
    \begin{cases}
    w_t - \eta  \left(\up{w} - w_* \right),   & \text{if }w \geq \bar{w}\\ 
    w_t - \eta  \left(\down{w} - w_* \right), & \text{if } w < \bar{w}\\
    \end{cases}$\\
    \\[-0.5em]
PSG  \citep{PBGS} & $ w_{t+1} = \begin{cases}
    w_t - \eta  \left(\up{w} - w_* \right)   \left(\up{w} -w + \epsilon\right),   & \text{if }w \geq \bar{w}\\
    w_t - \eta  \left(\down{w} - w_*\right) \left(w - \down{w}+ \epsilon\right),  & \text{if } w < \bar{w}\\
    \end{cases}$ \\
    \\[-0.5em]
  EWGS  \citep{EWGS} & $ w_{t+1} = \begin{cases}
    w_t - \eta  \left(\up{w} - w_* \right) \left(1+ \delta \left( w -\up{w}\right)\right), & \text{if }w \geq \bar{w}\\
    w_t - \eta  \left(\down{w} - w_*\right) \left(1- \delta \left(w - \down{w}\right)\right), & \text{if } w < \bar{w}\\
    \end{cases}$ \\
    \\[-0.5em]
  Dampening (sec. \ref{sec:oscillations_dampening}) & $ w_{t+1} = \begin{cases}
    w_t - \eta  \left(\up{w} - w_* + 2\lambda \left(w - \up{w}\right)\right), & \text{if }w \geq \bar{w}\\
    w_t - \eta  \left(\down{w} - w_* + 2\lambda \left(w - \down{w}\right)\right), & \text{if } w < \bar{w} \\
     \end{cases}$\\
       \\[-0.5em]
\end{tabular}
\end{center}

If we look at the equations for PSG and EWGS more carefully, we see that they introduce an \textit{always positive} scaling term to the STE gradient (assuming sufficiently small $\delta$) . Therefore, these methods can only change the magnitude of the update and not its direction, meaning they cannot stop oscillations from happening. Because the methods apply a  scaling to the STE gradient, we call them \textit{multiplicative} variations in section \ref{sec:related_work}. 

On the other hand, our dampening method adds an extra term to the gradient, which is of opposite sign to the STE gradient. Thus, this \textit{additive} variation of the STE can indeed prevent oscillations by changing the direction of the gradient.




\subsection{Relation of the distance and frequency}
\label{app:relation_distance_frequency}

\begin{figure*}[h]  
\centering
\includegraphics[width=0.9\textwidth]{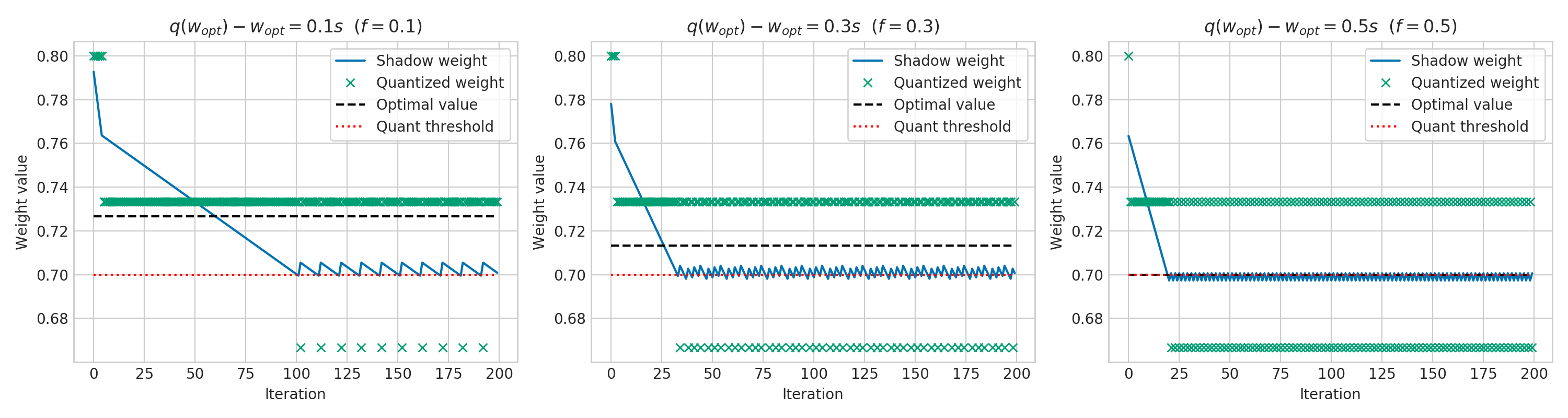}
\caption{Oscillation  of single weight in toy-regression problem for optimal values. The distance between the closest quantization point and the optimal value directly corresponds to the oscillation frequency.}
\label{fig:oscillation_frequency}
\end{figure*}

In toy example of section \ref{sec:oscillations}, we explained how the frequency of the oscillations depends linearly on distance of the optimal weight value $w_*$ from its closest quantization level $q(w_*)$. We can, thus, define the oscillation frequency as:
\begin{equation}
    f = \frac{|q(w_*) - w_*|}{s},
    \label{eq:toy_example_freq}
\end{equation}
where $s$ is the quantization scaling factor or size of the quantization bin. The frequency depends linearly on the distance $d= |q(w_*) - w_*|$ due to the squared error loss. In figure \ref{fig:oscillation_frequency}, we demonstrate this dependency experimentally by varying the distance $d$ and observing the effect on the oscillation frequency.

\subsection{Different learning rates and multiplicative regularization methods}
\label{app:learning_rates_oscillation}
\begin{figure*}[ht]  
\centering
\includegraphics[width=0.9\textwidth]{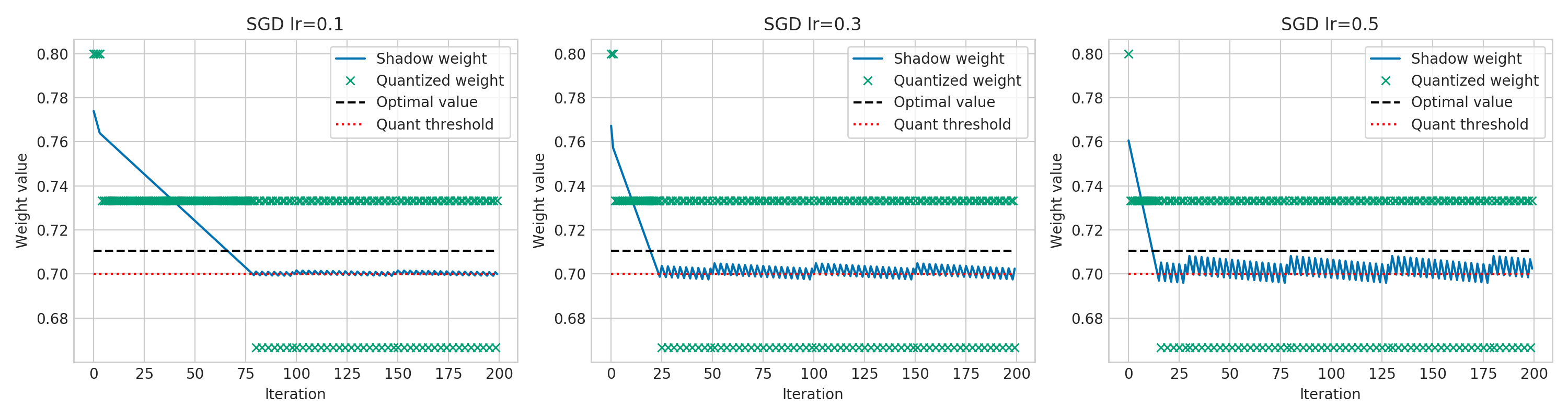}
\caption{Oscillation  of single weight in toy-regression problem for various learning rates (with STE).}
\label{fig:oscillation_lrs}
\end{figure*}

In figure \ref{fig:oscillation_lrs}, we plot the latent weight in our toy regression example for different learning rates assuming STE gradients. We can see that the change in learning rate only influences the amplitude of the oscillations and not the oscillation frequency. 

This is also apparent from equation \eqref{eq:toy_example_freq}, which shows that the frequency is independent of the learning rate. Theoretically, this suggests that \textit{multiplicative} variations of the STE, such as EWGS and PSG, cannot prevent oscillations from occurring.
Assuming a sufficiently small learning rate, weights will in the limit still converge to the decision threshold and start oscillating. While oscillating, the distance of the oscillating weight from its two closest quantization levels is roughly constant and equal to $s/2$.
This means that close to the decision threshold the weight update rules for PSG and EWSG become:
\begin{center}
\begin{tabular}{ l | l } 
PSG  \citep{PBGS} & $ w_{t+1} = \begin{cases}
     w_t - \eta  \left(\frac{s}{2}+ \epsilon\right) \left(\up{w} - w_*\right)  ,   & \text{if }w \geq \bar{w}\\
    w_t - \eta \left(\frac{s}{2} + \epsilon\right)  \left(\down{w}- w_* \right) ,  & \text{if } w < \bar{w}\\
    \end{cases}$ \\
    \\[-0.5em]
  EWGS  \citep{EWGS} & $ w_{t+1} = \begin{cases}
    w_t - \eta \left(1-  \frac{s\delta}{2}\right)  \left(\up{w} - w_*\right) , & \text{if }w \geq \bar{w}\\
    w_t - \eta \left(1-  \frac{s\delta}{2} \right)  \left(\down{w}-w_* \right) , & \text{if } w < \bar{w}\\
    \end{cases}$ \\
    \\[-0.5em]
\end{tabular}
\end{center}
Both methods now apply a constant and positive scaling to the gradient. Given that typically $s<1$, and that for EWGS $\delta<1$, the combined gradient scaling is $<1$ effectively reducing the learning rate and consequently the oscillation amplitude. This analysis is only valid if the oscillation amplitude is small, which is typical towards the end of training. However, in the earlier stages of training, where the weights move more freely, the dynamics of the oscillations might be more complicated, as the gradient or velocity of the weight depends on the distance from the closest quantization bin.

\subsection{The bias of the STE}
\label{app:biased_ste}
In the original paper on the straight-through estimator, \citet{bengio2013estimating} note that ``...\textit{(the STE) is clearly a biased estimator}''. Several authors, such as \citet{chmiel2022luq, fan2021extreme}, have repeated this statement and have argued that this \textit{bias} is a fundamental drawback of STE when applied to quantization, without defining what this actually means in the context of quantization-aware training.


The bias of the STE in the original formulation occurs in a probabilistic setting. Let's assume we want to optimize a loss $\mathcal{L} = \mathbb{E}_{p(z|\theta)} \left[f(z)\right]$ where $z$ is latent parameter coming for a discrete distribution $p(z|\theta)$. To learn the parameter $\theta$ we need to calculate the derivative $\nabla_{\theta} \mathbb{E}_{p(z|\theta)} \left[f(z)\right]$. However, due to the discrete nature of $p_z$ this derivative does not exist. Instead the STE provides us with the convenient approximation $\mathbb{E}_{p(z|\theta)}\left[\nabla_{z}f(z) \nabla_{\theta}\tilde{z} \right]$, where $\tilde{z}$ is the \textit{unquantized} version of $z$. In this case, indeed the STE is in most cases a biased estimator \cite{pervez2020lowbl}, as the expectation of the gradient does not match with the gradient of the expectation.


However, it is unclear what the bias of the estimator means in the QAT setting. Quantized networks do not have an inherent probabilistic property, unless explicitly defined so, e.g. in Bayesian neural networks or when training with stochastic rounding \cite{Gupta2015, louizos2018relaxed}. The gradients for the floating-point shadow weights are indeed \textit{wrong} or \textit{biased}, but these weights are by themselves meaningless because they are not used in the forward pass. As described in \cite{helwegen2019latent} the underlying weights of a neural network when trained with STE are not representative for the network's performance, but rather can be interpreted as a reservoir that slowly accumulates small gradient updates. Therefore, we believe that comparing to floating-point gradient in the conventional QAT setting can be misleading and nonconstructive.

\end{document}